# An optimized Capsule-LSTM model for facial expression recognition with video sequences

Siwei Liu[1]*, Yuanpeng Long[2]*, Gao Xu[1], Lijia Yang[1], Shimei Xu[3], Xiaoming Yao[1,3], Kunxian Shu[1#]

1. School of Computer Science and Technology, Chongqing Key Laboratory on Big Data for Bio Intelligence, Chongqing University of Posts and Telecommunications, Chongqing 400065 China
2. School of Economic Information Engineering, Southwestern University of Finance and Economics, Chengdu 611130, China
3. 51yunjian.com, Hetie International Square, Building 2 Room404, Chengdu, Sichuan, China

\* **Equal contribution**

\# **Corresponding author:** Kunxian Shu (e-mail: shukx@cqupt.edu.cn).

**Abstract**

To overcome the limitations of convolutional neural network in the process of facial expression recognition, a facial expression recognition model Capsule-LSTM based on video frame sequence is proposed. This model is composed of three networks includingcapsule encoders, capsule decoders and LSTM network. The capsule encoder extracts the spatial information of facial expressions in video frames. Capsule decoder reconstructs the images to optimize the network. LSTM extracts the temporal information between video frames and analyzes the differences in expression changes between frames. The experimental results from the MMI dataset show that the Capsule-LSTM model proposed in this paper can effectively improve the accuracy of video expression recognition.

**Key words** Facial expression recognition; Capsule-LSTM; Video frame sequence; Deep learning

## 1. Introuduction

Facial expression recognition has a wide range of application scenarios, and it has played an important role in fatigue driving recognition, human-computer interaction, clinical medicine and other fields [1, 2]. The process of facial expression recognition is roughly divided into three parts: preprocessing, feature extraction and recognition classification. Feature extraction algorithms can be divided into traditional machine learning methods [3, 4, 5] and deep learning methods. When dealing with more complex situations such as different lighting, postures, and occlusions, deep learning method is more applicable compared totraditional methods. Most of the current mainstream facial expression recognition algorithms are implemented using deep learning method. In 2015, Sun et al. [6] used R-CNN to study facial expression features, extracted MSDF, DCNN and RCNN features, and trained linear support vector machine classifiers for these features. They have proposed a new fusion network to combine all the extracted features at the decision level. Li et al. [7] proposed a facial expression recognition method based on Faster R-CNN. Hamester et al. [8] proposed a facial expression recognition method based on multi-channel convolutional neural network. In the work of Zhou et al. [9], a multi-dimensional facial expression recognition method was proposed, and the bilinear pool method was used to encode the second-order statistics of the features.

With the ongoing improvement of computing power and algorithms, the current facial expression recognition research has gradually shifted from the study of static pictures to the video recognition of dynamic scenes. More information can be gained from the temporal correlation of consecutive frames..Yacoob et al. [10] obtained the temporal and spatial representations of consecutive frames in the optical flow and gradient domain. Sanchez et al. [11] used optical flow method to locate and track facial key points for facial expression recognition. Pantic et al. [12] screened out 15 key points in FACS, tracked them, and analyzed the trajectories for facial expression recognition. Zhang et al. [13] proposed a hierarchical two-way recurrent neural network to analyze facial expression information in temporal series, extracting sequential features from continuously changing face key points.  Their work hasalso proposed a multi-signal convolutional neural network to extract spatial features from static frames.

## 2. Related Work
### 2.1 Capsule

CNN has become one of the most popular deep learning networks and is commonly utilized in the field of image recognition. However, CNN's ability to achieve consistency after image migration is low [14]. The image migration here means that the convolutional neural network has difficulty to detect the consistency of the left and right translation and rotation. CNN has good performance in object feature extraction and detection, but it ignores local and internal relative position information (such as relative position, direction, skewness, etc.), thus losing some important information.

In order to improve CNN, Hinton et al. [15] proposed a simple three-layer Capsule network. Unlike CNN, when Capsule recognizes a specific object in a certain area of the image, the output is a capsule vector. It is no longer a scalar. The modulus length of the vector represents the possibility of identifying the object. The direction represents the internal information contained in the object, such as relative position and orientation. If the object is slightly changed, such as rotation, translation, orientation, size, etc., the output vector modulus length will not change. However, the direction will be slightly changed accordingly. This allows Capsule to use a simple and unified architecture to deal with different visual tasks.

The Capsule network includes two parts: encoder network and decoder network. The function of the encoder is similar to that of the convolutional neural network, which is to map the input image from the bottom up to the feature space. The Capsule encoder network is mainly composed of three parts: convolutional layer, Primary Caps and digital capsule layer. The network structure is shown in Figure 2.1.

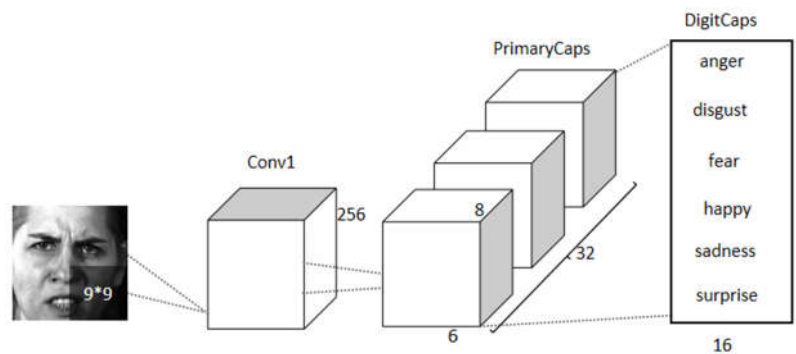

Figure 2.1 Capsule encoder structure diagram

The Capsule decoder optimizes the network by constructing reconstruction loss. The decoder network consists of three fully connected layers and is a simple feedforward network. The Capsule decoder only retains the capsule vector corresponding to the prediction category of the Capsule encoder, and reconstructs the image through the capsule vector. The Capsule decoder network structure is shown in Figure 2.2.

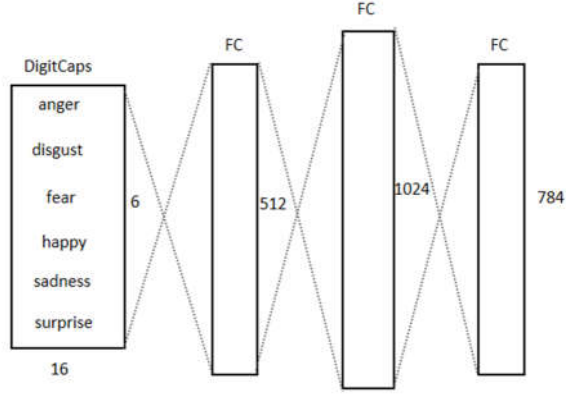

Figure 2.2 Capsule decoder structure diagram

## 2.2 Dynamic routing algorithm

The dynamic routing algorithm is the core of the Capsule network [15]. The low-level capsule uses a dynamic routing iterative algorithm to predict high-level features. The bottom-layer capsule is obtained through the convolutional layer. To obtain the high-layer capsule, the bottom-layer capsule needs to be calculated by formula 2.1 and 2.2. The calculation process of these two formulas is to multiply the weight and sum.

$$S_j = \sum_i c_{ij} \hat{u}_{j|i} \tag{2.1}$$

$$\hat{u}_{j|i} = W_{ij} u_i \tag{2.2}$$

Where $\hat{u}_{j|i}$, $u_i$ are the vector of the capsule, $w_{ij}$ is the weight parameter, and $C_{ij}$ is the coupling coefficient. Each capsule in the upper layer and each capsule in the bottom layer in the model has a coefficient. The sum of these coefficients is 1, This coefficient is calculated by the formula below.

$$b'_{ij} = b_{ij} + v_j \hat{u}_{j|i} \tag{2.3}$$

$$c_{ij} = soft\max(b'_{ij}) \tag{2.4}$$

$b_{ij}$ is the logarithmic prior probability, and is initially zero. $v_j$ is the output vector of the capsule. In the iterative process, $b_{ij}$ is continuously updated according to the consistency between the output capsule vector $v_j$ of the current layer and the prediction vector $\hat{u}_{j|i}$ of the previous layer. The consistency is determined by the d scalar product of $c_{ij}$ and $b_{ij}$.

### 2.3 Research status of capsule network

Capsule has demonstrated its unique advantages forimage recognition,text generation [16] and other applicantions. Yu [17] et al. used the capsule network for finger vein recognition applications, and has improved the recognition accuracy by combining the existing CNN model with the capsule network structure. Hollósi et al. [18] selected three neural networks, VGG, Res Net and Dense Net, and improved the recognition accuracy of the neural network by adding capsule layers. In addition, the capsule network also has a good application on small-scale data sets processing[19].

## 3. Methodology

### 3.1 Network Model

The Capsule-LSTM model is shown in Figure 3.1.This network is mainly composed of three parts: Capsule encoder, Capsule decoder and LSTM. A picture with a size of 48*48 is passed through a convolutional layer with a convolution kernel of 3, a convolutional capsule layer, and a digital capsule layer of the Capsule encoder to obtain N feature vectors of length 30, where N is the number of labels. On this basis, only the feature vector with the largest modulus length is selected and input to the Capsule decoder. The decoder uses the deconvolutional layer network reconstruction network proposed by other researchers [20] to calculate the Euclidean distance pixel by pixel from the reconstructed 48*48 image and the original image. The LSTM network is used to extract the temporal information between consecutive video frames. After the image passes through the Capsule encoder, the N*30 feature matrix is obtained. Softmax function is then used to convert the feature matrix into N*1.The time interval of LSTM is set to 16, and 16 consecutive frames are selected in each video. The input of the LSTM network is 16*N*1. The number of hidden layers of the LSTM network is set to 128 in this network.The value with the largest expression probability is selected as the predicted value of the video sequence model to complete the video expression classification of the sequence samples.

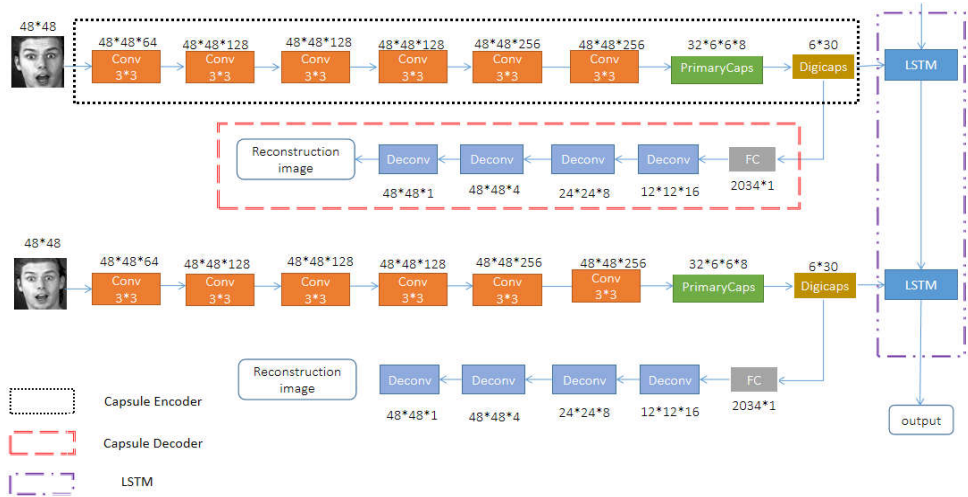

Figure 3.1 Capsule-LSTM network structure diagram

A human face is a structured entity, which contains local features such as eyes, nose, mouth, facial muscles, hair, and beard. The local features have certain arrangement rules and relative spatial information. In the Capsule decoder, the convolutional layer of the multi-layer small convolution kernel is used in series, which can reduce the amount of model parameters and ensure the receptive field of the convolutional layer. Combined with the dynamic routing algorithm, the internal structural features of the face can be effectively extracted, thereby improving the ability of spatial feature extraction. Capsule reconstruction network can further optimize the network. In order to extract the temporal characteristics between frames, the LSTM network is introduced, which can better solve the long-term dependency problem compared to RNN. The model porposed in our study can effectively conduct facial expression recognition in video by combining Capsule and LSTM.

**3.2 Loss Function**

For the Capsule-LSTM network, this paper defines a joint loss function, which consists of three parts, as shown in formula 3.1:

$$total\_loss = margin\_loss + reconstruction\_loss + Lstm\_Loss \quad (3.1)$$

The joint loss function is composed of marginal loss, reconstruction loss, and cross-entropy function. The Capsule encoder uses the marginal loss to realize multi-class recognition. The capsule corresponding to each class object uses the marginal loss function to obtain the class loss function. The total marginal loss is the sum of the loss of all expression classes. The marginal loss is shown in formula 3.2.

$$margin\_loss = T_c \max(0, m^+ - \|v_c\|)^2 + \lambda (1-T_c) \max(0, \|v_c\| - m^-)^2 \quad (3.2)$$

The value of $T_C$ is only 0 or 1. Expression category c exists as 1, and does not exist as 0. $m^+$ represents the upper boundary value, which is 0.9. $m^-$ represents the lower boundary value, which is 0.1. $\|v_c\|$ is expressed as the module length of the capsule, which is expressed as the probability. The probability may be an expression category. The value of $\lambda$ is 0.5, which is used to reduce the loss caused by non-existent classes. Each class uses a separate marginal loss, and the total loss is the sum of the marginal losses of all expression classes.

In order to enable the capsule to extract features from the input image and output corresponding instantiation parameters, a reconstruction loss is used as a regular term asshown in formula 3.3.

$$reconstruction\_loss = 0.0005 * \frac{1}{n} \sum_{i=1}^{n} (r_i - a_i)^2 \quad (3.3)$$

Among them, $r_i$ represents the number of pixels in the original image, and $a_i$ can be regarded as the reconstruction value of the i-th pixel. Because the calculated value of reconstruction loss is very large, in order to avoid reconstruction loss dominate the trend of the entire joint loss function, our model multiplies the reconstruction loss value by a very small value, which is 0.0005.

In the LSTM network, cross entropy is used as the loss function, as shown in formula 3.4.

$$Lstm\_Loss = 0.5 * - \sum_i y_i' \log(y_i) \quad (3.4)$$

Where $y_i'$ is the actual expression category label, and $y_i$ is the expression probability predicted by sample i. The task of this paper is multi-classification, and the calculated value of cross entropy is large. The cross entropy function is multiplied by 0.5 to balance the ratio between the loss function, the marginal function and reconstruction function of the LSTM network.

## 4. Experiments and Results

### 4.1 Data set and data processing

The MMI facial expression database [21] contains more than 2900 expression videos of more than 75 subjects. It contains not only the expressions of the six basic emotions (angry, disgust, fear, happiness, sadness, surprise), but also the expressions with FACS action unit marks including basic action units and other action identifiers. This paper selects 208 video sequences that have been marked with six basic expression labels in the MMI data set, and all the data in the MMI data set are videos. The video is converted into video frames using ffmpeg. We used the image face detection library Dlib to perform face detection, divided the face part, and selected the middle 16 frames of peak expressions as the video sequence. One video sequence corresponds to one label. The selected video frame was processed into a grayscale image with a size of 48*48. All images were flipped horizontally to obtain mirror image data, and each image was rotated by 5°, 10°, 15°, -5°, -10°, -15° degrees. The final data volume is 8 times that of the original data set.

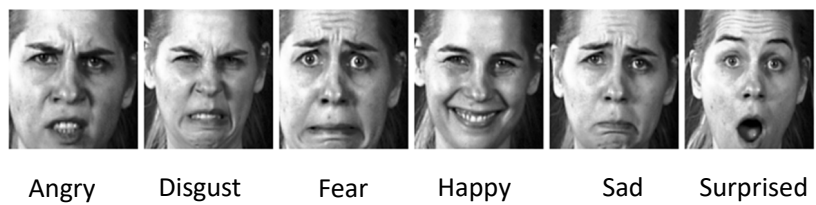

Angry　　Disgust　　Fear　　Happy　　Sad　　Surprised

Figure 4.1 MMI data set

**4.2 Experiment Settings**

The computer hardware conditions and configuration used in the experiment in this article are as below: the processor is I7-8700, the GPU is GTX 2080Ti, the operating system is Ubuntu 18.04, and the development language used in this article is Python 3.7. In order to improve the efficiency of model training, a deep learning framework based on TensorFlow-GPU is used. In the training process of the network, the Adam optimizer is used to optimize the network, and the learning rate is set to 0.0001. This article uses classification accuracy and confusion function as model evaluation criteria.

**4.3 Experimental Results**

During the experiment, a variety of loss functions were designed for the Capsule-LSTM network model proposed in this paper, and verified using the MMI data set. The specific experimental results obtained are shown in Table 4.1. Experiments have been carried out that the Capsule encoder uses the marginal loss function, the Capsule decoder uses the reconstruction loss function, and the LSTM uses the cross-entropy function. The experiments have showed that the best result is 72.11% by using the combined loss function of these three loss functions.

Table 4.1 Comparison of the results of various loss functions with the MMI data set

| Loss Function | ACC(%) |
| --- | --- |
| margin_loss（Capsule）+ margin_loss（LSTM） | 65.46 |
| margin_loss（Capsule）+reconstruction_loss（Capsule）+ margin_loss（LSTM） | 68.30 |
| margin_loss（Capsule）+reconstruction_loss（Capsule）+ Cross Entropy（LSTM） | **72.11** |
| margin_loss（Capsule）+ Cross Entropy（LSTM） | 69.50 |

Figure 4.2(a) shows the curve of the margin loss of the Capsule encoder. Figure 4.2(b) shows the curve of the reconstruction loss of the Capsule decoder. Figure 4.2(c) shows the change trend of LSTM using cross entropy. Figure 4.2(d) shows the change trend of Capsule-LSTM using the joint loss function on the data set MMI. The joint loss function is the weighted sum of the three loss functions.

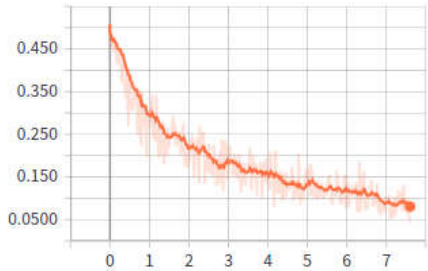
4.2(a) Trend of the Margin Loss

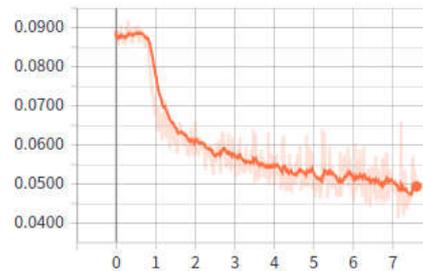
4.2(b) Trend of the reconstruction Loss

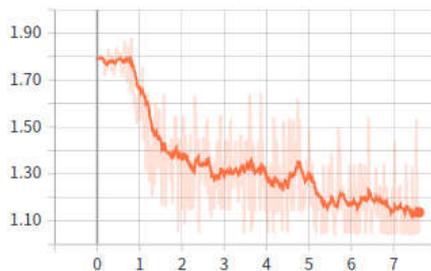
4.2(c) Trend of the Cross Entropy

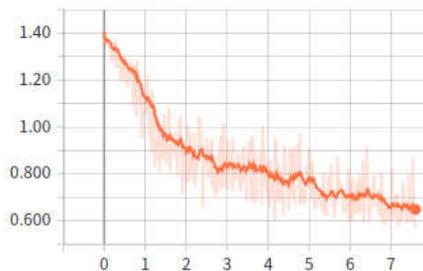
4.2(d) Trend of the joint loss function

Figure 4.2 The downward trend of the loss function in the MMI data set

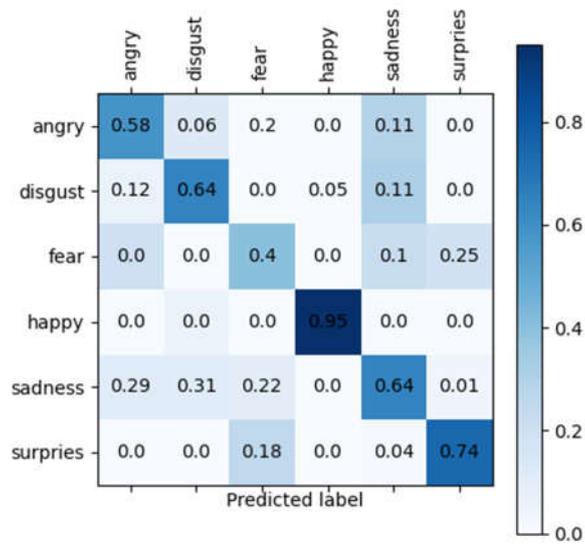

Figure 4.3 The confusion matrix of Capsule-LSTM on the MMI dataset

Figure 4.3 shows the result of the loss function on the MMI data set during the experiment. It can be seen from the figure that the ability to recognize happy and surprised expressions is better. The recognition rate of happy expression reaches 95%, and the characteristics of happy expressions are more obvious. Therefore, the recognition accuracy rate is higher. However, the recognition of the two emotions of fear and anger on the MMI data set is relatively poor. The recognition accuracy of the fear emotion is below 50%. The sample data of the two emotions of fear and anger is relatively small, and the salient features of facial expressions are not obvious enough. It is easy to be misjudged as sadness.

In order to further verify the effectiveness of Capsule-LSTM, the experimental results are compared with other network models, including 3DCNN[22], IL-CDD[23], CSACNN[24], DTGAN[25]. The experimental results are shown in Table 4.2.

Table 4.2 Comparison with other algorithms

| Model | ACC(%) |
|---|---|
| 3DCNN[22] | 60.45 |
| IL-CDD[23] | 65.69 |
| CSACNN[24] | 71.52 |
| DTGAN[25] | 69.53 |
| Capsule-LSTM | **72.11** |

## 5. Conclusions

This paper proposes a better network model based on Capsule-LSTM for video expression recognition. Capsule network is utilized for spatial feature information extraction, and LSTM network is used for processing continuous video frame information. For the Capule-LSTM network model, a joint loss function is proposed. Experiments show that it can better improve the performance of the network. For the Capsule-LSTM network model proposed in this paper, experiments are carried out on the public data set MMI. Based onthe network model proposed in this paper, follow-up research can use more complex AU (Face Motion Unit) data sets to design a more robust network.